\documentclass[letterpaper, 10 pt, conference]{IEEEtran}  

\IEEEoverridecommandlockouts                              

\usepackage[pdftex]{graphicx}
\graphicspath{{figures/}}
\usepackage{hyperref}
\hypersetup{
    colorlinks=true,
    linkcolor=black,
    citecolor=black,
    filecolor=black,
    urlcolor=black,
}
\usepackage[T1]{fontenc}
\usepackage[utf8]{inputenc}
\usepackage[english]{babel}
\usepackage[export]{adjustbox}
\usepackage{caption}
\usepackage{subfig} 
\usepackage[colorinlistoftodos]{todonotes}
\usepackage{lipsum}
\usepackage[]{algorithm2e}
\usepackage{amsmath}

\usepackage{placeins}

\title{\LARGE \bf
LIMO\@: Lidar-Monocular Visual Odometry
}

\author{Johannes Graeter$^{1}$, Alexander Wilczynski$^{1}$ and Martin Lauer$^{1}$
\thanks{$^{1}$Institute of Measurement and Control,
        Karlsruhe Institute of Technology, Germany.
        {\tt\small johannes.graeter@kit.edu}}%
}

\begin{document}

\maketitle

 \IEEEpubid{\begin{minipage}{\textwidth}~\\[12pt] \centering%
   10.1000/xyz123~ 
   \copyright~2018 IEEE. Personal use of this material is permitted. Permission from IEEE must be obtained for all other uses, including reprinting/republishing this material for advertising or promotional purposes, collecting new collected works for resale or redistribution to servers or lists, or reuse of any copyrighted component of this work in other works.
 \end{minipage}}
 \IEEEpubidadjcol

\pagestyle{empty}

\begin{abstract}
Higher level functionality in autonomous driving depends strongly on a precise motion estimate of the vehicle.
Powerful algorithms have been developed. However, their great majority focuses on either binocular imagery or pure LIDAR measurements.
The promising combination of camera and LIDAR for visual localization has mostly been unattended.
In this work we fill this gap, by proposing a depth extraction algorithm from LIDAR measurements for camera feature tracks and estimating motion by robustified keyframe based Bundle Adjustment.
Semantic labeling is used for outlier rejection and weighting of vegetation landmarks.\\
The capability of this sensor combination is demonstrated on the competitive KITTI dataset, achieving a placement among the top 15.
The code is released to the community.
\end{abstract}

\section{Introduction and Related Work}
\label{sec:introduction}
\begin{figure}[t]
  \small
  \centering
  \def\svgwidth{\columnwidth}
  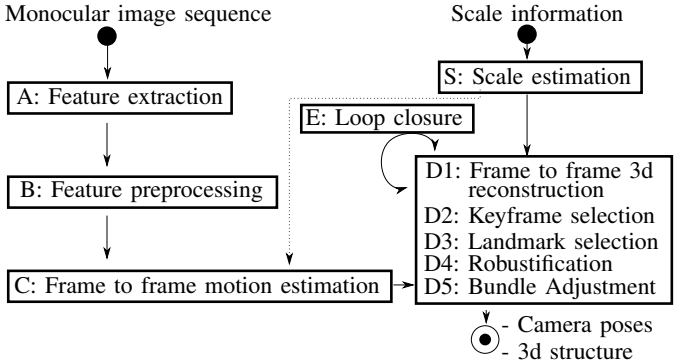%
  \caption{Structure of the VSLAM pipeline. 
  The input is a temporal image sequence, output are the camera poses and a sparse reconstruction of the environment.
  Scale information can be retained from a second, calibrated camera with known baseline~\cite{mur2017orb}, structure inherent information, such as the distance to the ground plane~\cite{grater2015robust}, IMU~\cite{nutzi2011fusion} or in our case depth measurements from a LIDAR\@.
  Scale can optionally be used for estimating the frame to frame motion as well.
  Note that in this work, we aim for Visual Odometry, hence no loop closure is performed.
  }
\label{fig:organigram_global}
\end{figure}

Even though Structure from Motion algorithms have a history over nearly 100 years~\cite{cremers2017direct}, it is still subject to research.
Today often being revered to as Visual Simultaneous Localization and Mapping (VSLAM) or Visual Odometry, depending on the context (see~\cite{taketomi2017visual}), the basic idea is a simple one --- by observing the environment with a camera, its 3d structure and the motion of the camera are estimated simultaneously.
The most popular method for VSLAM is called Bundle Adjustment, since it aligns bundles of lines of sight focus in observed points, called landmarks. 
Having all landmarks and camera poses as parameters, this optimization problem can become large and costly to solve.
Recent development in computer hardware and digital imagery enables the use of offline VSLAM for mapping and localization (~\cite{hartley2003multiple},~\cite{szeliski2010computer},~\cite{sons2015multi}).
However, in order to use VSLAM for online mapping and trajectory estimation, the complexity of the optimization problem has to be reduced, which is a recent field of research.\\
\begin{figure}[t]
  \small
  \centering
  \def\svgwidth{\columnwidth}
  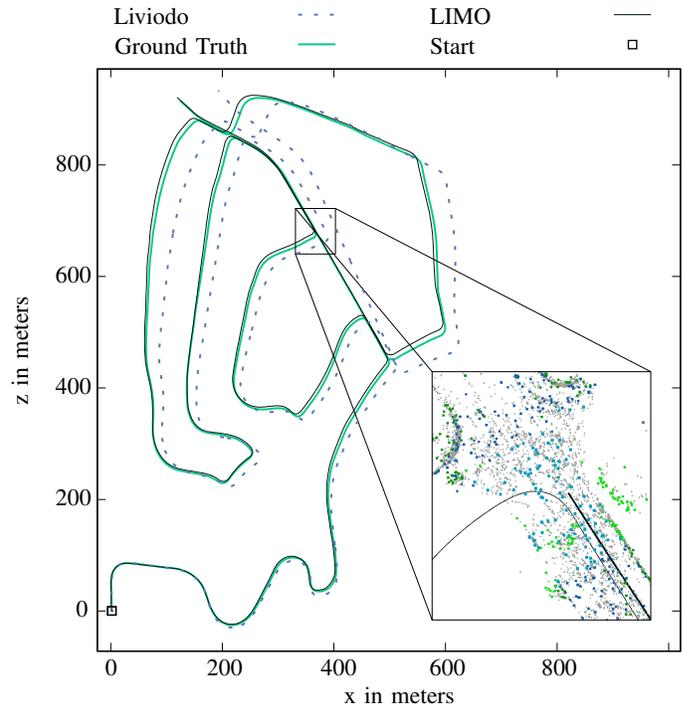%
  \caption{Trajectory of the frame to frame motion estimation (Liviodo; blue, light dashed) and the motion estimation with keyframe Bundle Adjustment (LIMO\@; black, solid) of sequence 02 from the KITTI benchmark. 
  The estimated trajectory can trace the ground truth (orange, solid) with very high accuracy and low drift without loop closing.
  In zoom, we show landmarks from monocular Visual Odometry (near: cyan, middle: blue) and landmarks with estimated depth from LIDAR (near: light green, middle: dark green), which are used in LIMO for Bundle Adjustment.
  Rejected landmarks are shown in gray.
  }
\label{fig:intro}
\end{figure}
In order to exemplify the different approaches in VSLAM, we illustrate its different building blocks in Figure~\ref{fig:organigram_global}.
Since stereo matching is a well explored field of research, the majority of methods obtain scale information (Block S) from a second camera (\cite{mur2017orb},~\cite{sons2015multi},~\cite{geiger2011stereoscan},~\cite{cvivsic2017soft}).
The most recent top performing algorithms such as ROCC~\cite{buczko2016flow} and SOFT~\cite{cvivsic2015stereo} focus on feature extraction and preprocessing (Block A and B), extracting precise and robust feature tracks and selecting them with sophisticated techniques. 
Even without Bundle Adjustment (Block D) they achieve very good results on the challenging KITTI Benchmark~\cite{geiger2013vision}.
Adding block D to SOFT results in the top performing binocular Visual Odometry method on KITTI~\cite{cvivsic2017soft}.\\
The main drawback of stereo vision is its strong dependency on a precise extrinsic camera calibration.
Krevso et al.~\cite{krevso2015improving} found, that learning a compensation of the calibration bias through a deformation field improves the performance drastically.
\IEEEpubidadjcol
The top performing stereo algorithm SOFT-SLAM also uses a calibration error compensation in the translation estimation, that was originally introduced by Geiger et al.~\cite{geiger2011stereoscan}.
This is not a matter of careless calibration --- since already small errors in the calibrated camera pose can lead to large errors in the estimated depth, the preservation of a correct stereo camera calibration is difficult and costly.\\
The error of depth measurements from a LIght Detection And Ranging sensor (LIDAR), however, is small and independent of the extrinsic calibration error.
We aim to combine the best of both worlds, the highly accurate depth estimation from LIDAR and the powerful feature tracking capability of the camera.
Since the LIDAR measures in a different domain, an additional step has to be added that combines feature tracking and the depth measured by the LIDAR\@.
Moreover, LIDAR to camera calibration is still an active field of research (\cite{geiger2012automatic},\cite{grater2016photometric}) and its accuracy is limited to a few pixels. 
As a result, VSLAM with LIDAR and a monocular camera is an unattended topic, mainly covered by Zhang et al.~\cite{zhang2014real},~\cite{zhang2015visual} and Caselitz et al.~\cite{caselitz2016monocular}.\\
Since that sensor combination shows potential for further improvement, we contribute a new method for extracting depth from LIDAR for feature points in an image as described in Section~\ref{sec:depth_estimation}. 
We reject outliers that do not satisfy the local plane assumption and treat points on the ground plane specially for greater robustness. 
The depth information is fused with prevalent monocular techniques in the VSLAM pipeline presented in Figure~\ref{fig:organigram_global}.
In that way, we can take advantage of data accumulation and temporal inference to lower drift and increase robustness (Section~\ref{sec:backend}).
We aim for Visual Odometry and do therefore not employ loop closure.
In order to fulfill real time constraints, we take special care of the prior estimation, landmark selection and keyframe selection.   
In this work, we explicitly do not use any LIDAR-SLAM algorithmic such as ICP, used by Zhang et al.~\cite{zhang2015visual} in order to push the limits of the combination Visual Odometry and depth information.  
The presented methodology is evaluated on the competitive KITTI benchmark, being ranked 13th\footnote{As of 1st of March 2018.} in terms of translation error and 11th in terms of rotation error, outperforming state of the art methods such as ORB-SLAM2~\cite{mur2017orb} and Stereo LSD-SLAM~\cite{engel2015large}.

\section{Block A and B:\@ Feature extraction and preprocessing}
The feature extraction (Block A) as shown in Figure~\ref{fig:organigram_global} consists of a feature tracking step and a feature association step.
We employ the feature tracking methodology, implemented in the viso2 library~\cite{geiger2011stereoscan}.
It comprises non-maximum suppression, outlier rejection by flow and subpixel refinement.
Since the custom descriptor of viso2 is fast to extract and compare, 2000 feature correspondences can be computed in $30-40 \text{ms}$.\\
Landmarks lying on moving objects are particularly bothersome, in particular if only few feature points are available such as in highway scenarios.
Therefore, we lever the power of deep learning to reject landmarks lying on moving objects.
In the image domain, we scan the surroundings of the feature point in a semantic image~\cite{cordts2016cityscapes}.
If the majority of neighboring pixels belongs to a dynamic class, like vehicle or pedestrian, the landmark is excluded.

\section{Block S:\@ Scale Estimation}
\label{sec:depth_estimation}
In order to estimate scale (Block S), depth corresponding to detected feature points is extracted from the LIDAR\@.
In this work, we use a one shot depth estimation approach.
Whereas this results in less data available for feature depth estimation as using accumulated point clouds, we hence avoid the dependency to a motion estimate.\\

\subsection{Approach}
\label{sec:depth_estimation:approach}



First the LIDAR point cloud is transformed into the camera frame and projected onto the image plane.
For each feature point $f$ in the image the following steps are executed:
\begin{enumerate}
  \item Choose a set $F$ of projected LIDAR points in a given region of interest around $f$, see Section~\ref{sec:depth_estimation:roi}.
  \item Create a new set $F_{\text{seg}}$ by segmenting the elements of $F$ that are in the foreground by the method described in Section~\ref{sec:depth_estimation:segmentation}.
  \item Fit a plane $p$ to the elements in $F_{\text{seg}}$ with the method described in Section~\ref{sec:depth_estimation:plane_fit}.
        If $f$ belongs to the ground plane, a special fitting algorithm is used which is elaborated in Section~\ref{sec:depth_estimation:special_case_gp}.
  \item Intersect $p$ with the line of sight corresponding to $f$ to get its depth.
  \item Perform a test for the estimated depth.
  To be accepted as a depth estimate, the angle between the line of sight of the feature point and the normal of the plane must be smaller than a threshold.
  Moreover, we consider depth estimates of more than $30 \text{m}$ as uncertain and reject them.
\end{enumerate}

\subsection{Selecting the Neighborhood}
\label{sec:depth_estimation:roi}
\begin{figure}[t]
  \centering
  \includegraphics[width=0.8\columnwidth]{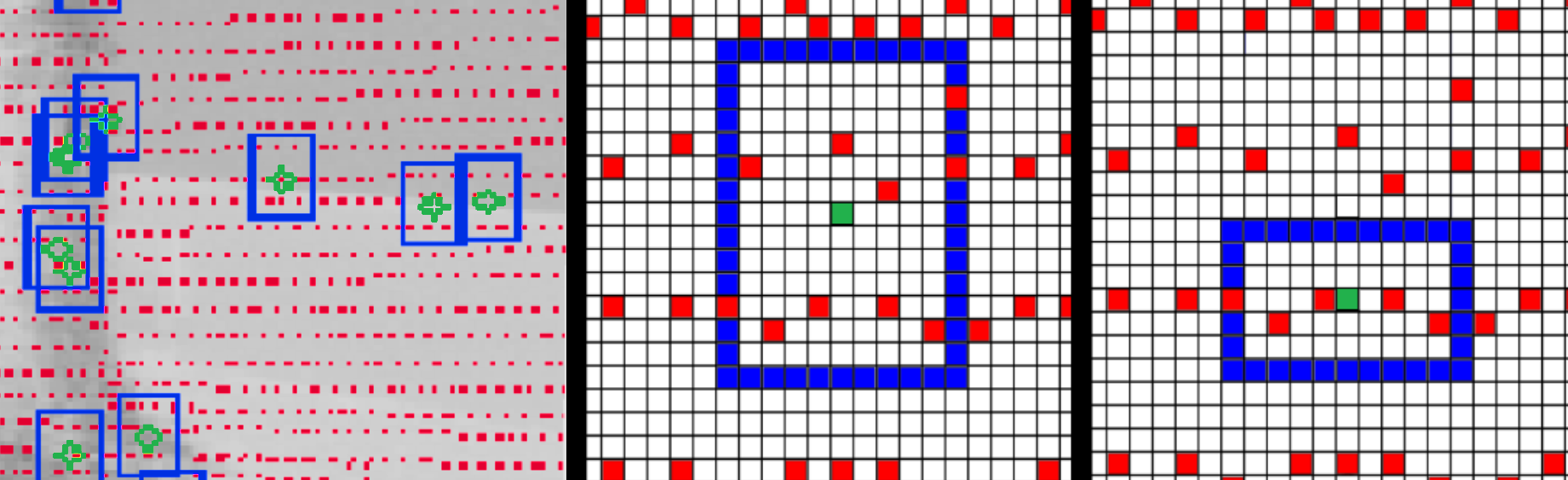}
  \caption{
  Choosing an adequate region of interest (blue) for determining the neighborhood of a detected feature point (green) in the projected point cloud (red) as shown in the left is important.
  The region of interest must include LIDAR points on a plane (middle) and not only on a line (right).}
\label{fig:roi}
\end{figure}

In order to extract the local plane around the feature point $f$, the first step is the selection of a neighborhood.
For ordered point clouds this step is trivial, since the neighborhood can directly be extracted from the point cloud.
For unordered point clouds, we use the projections of the LIDAR points on the image and select points within a rectangle in the image plane around $f$.
The rectangle size must be chosen such as singularities in plane estimation are avoided, as illustrated in Figure~\ref{fig:roi}.

\subsection{Foreground Segmentation}
\label{sec:depth_estimation:segmentation}
\begin{figure}[htb]
  \centering
  \subfloat[ The points on the background tilt the estimated local plane, causing a wrong depth estimate.]{\includegraphics[width=0.49\columnwidth]{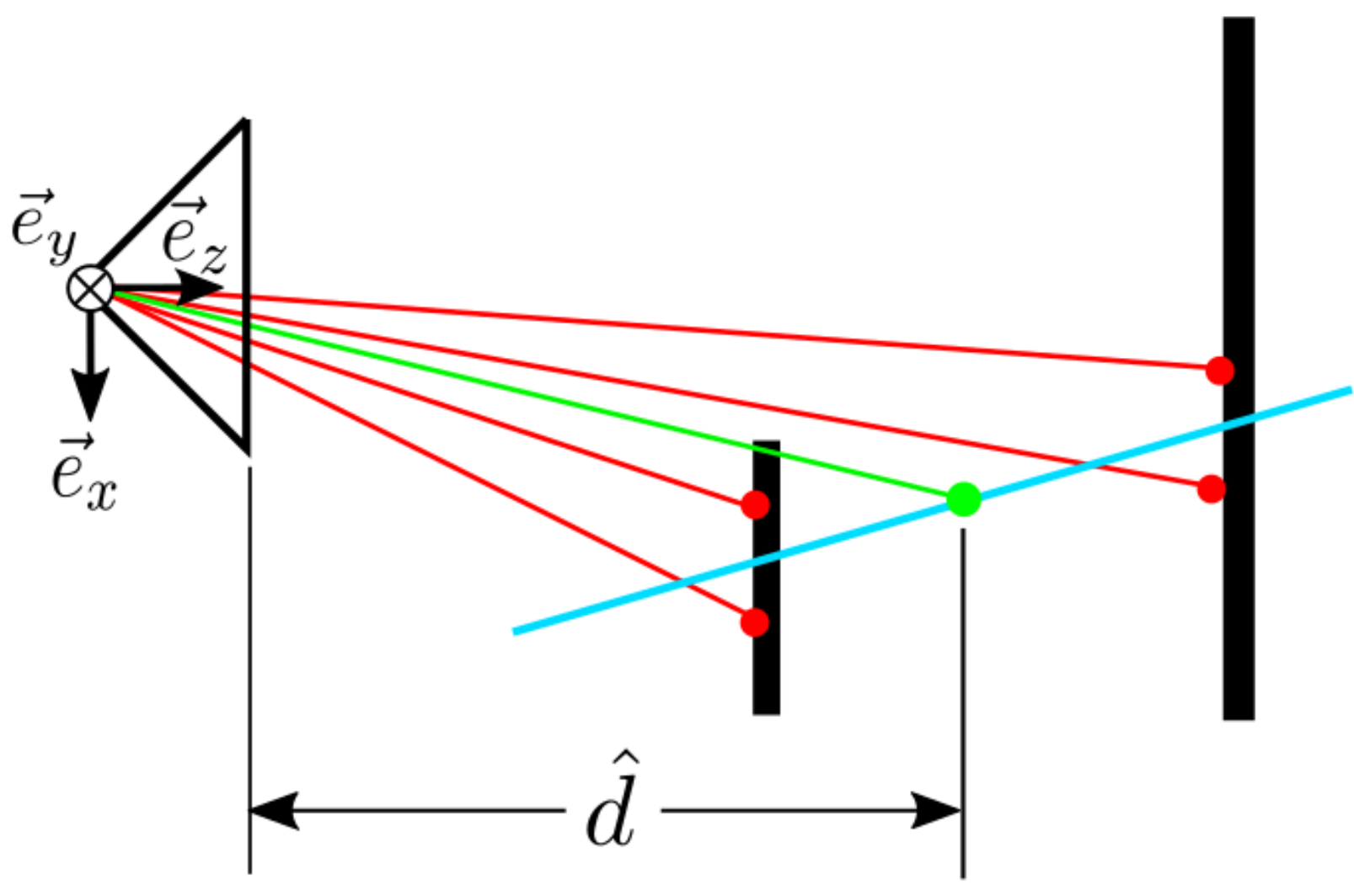}%
\label{fig:plane_fit_raw}}
  \hfill
  \subfloat[Only the points in the foreground are used for the local plane estimation resulting in a correct depth estimate.]{\includegraphics[width=0.49\columnwidth]{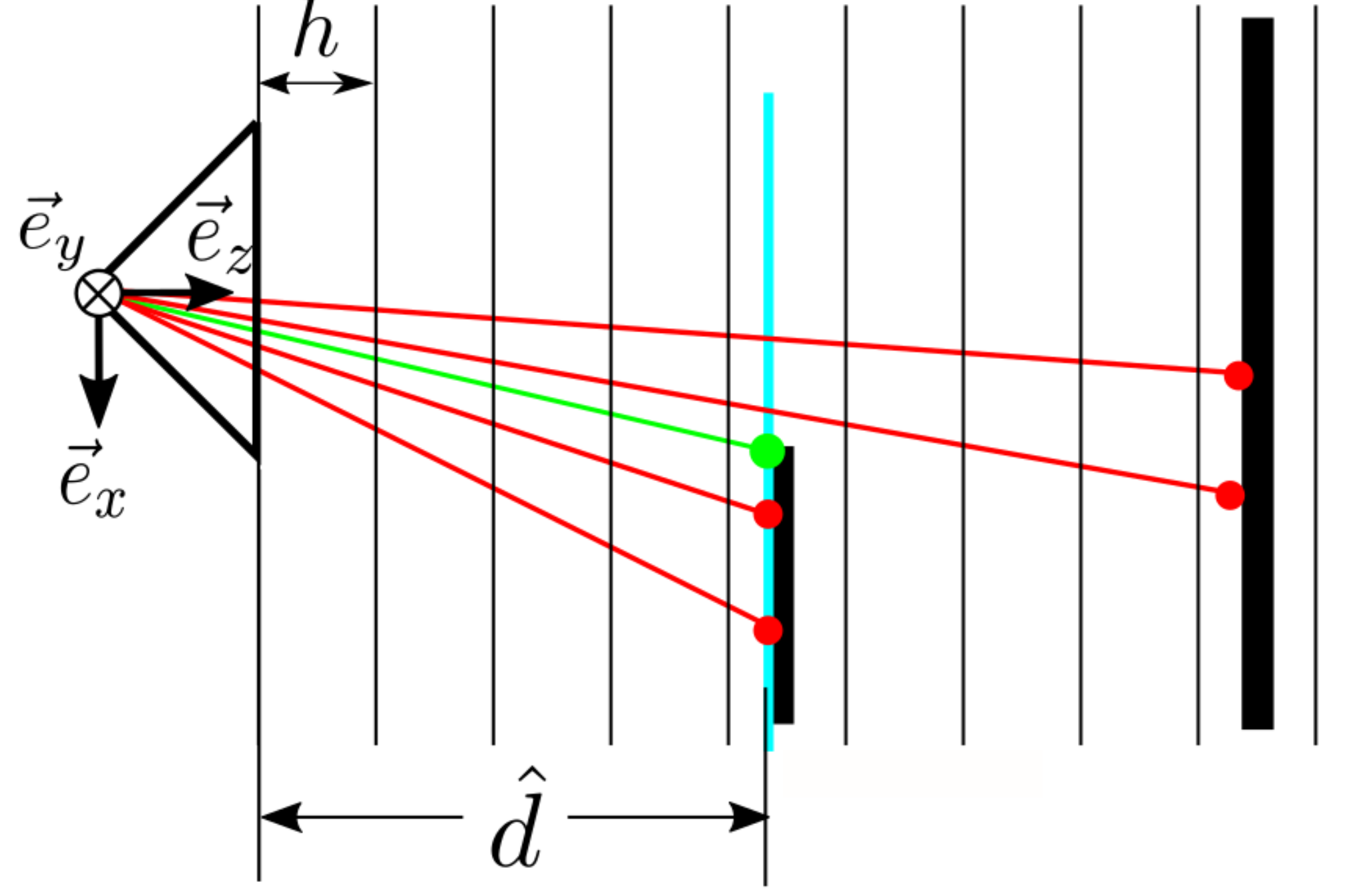}%
\label{fig:plane_fit}}
\caption{Depth estimation without segmentation (a) and with segmentation (b) by a depth histogram with bin width $h$.}
\end{figure}

If the feature point $f$ lies on a flat surface such as facades, the plane assumption is satisfied and the geometry around $f$ can accurately be approximated by the local plane fit through the set of LIDAR points in its vicinity $F$.
However, feature points usually lie on edges or corners and not on plane surfaces, since corners are better to track in image sequences (see~\cite{shi1994good}).
Fitting a plane to $F$ would therefore often lead to wrongly estimated depth as illustrated in Figure~\ref{fig:plane_fit_raw}.

To counter that, we segment the foreground $F_{\text{seg}}$ before executing the plane estimation.
Elements in $F$ are inserted into a histogram by depth with fix bin width of $h=0.3\text{m}$, see Figure~\ref{fig:plane_fit}.
A jump in depth from foreground to background corresponds to a gap in bin occupancy.
By choosing the LIDAR points of the nearest significant bin, segmentation can be employed efficiently for all detected feature points.
Since $f$ is tracked as an edge of the foreground plane, fitting the plane to $F_{\text{seg}}$ can accurately estimate the local surface around $f$.

\subsection{Plane Fit}
\label{sec:depth_estimation:plane_fit}
From the points in $F_{\text{seg}}$ we choose three points that span the triangle $F_{\Delta}$ with maximum area to stabilize the estimation.
If the area of $F_{\Delta}$ is too small, we do not use the depth estimation to avoid wrongly estimated depth.
We then fit a plane to $F_{\Delta}$, which is used for depth estimation.

\subsection{Special Case: Points on Ground Plane}
\label{sec:depth_estimation:special_case_gp}
\begin{figure}[t]
  \centering
  \includegraphics[width=0.7\columnwidth]{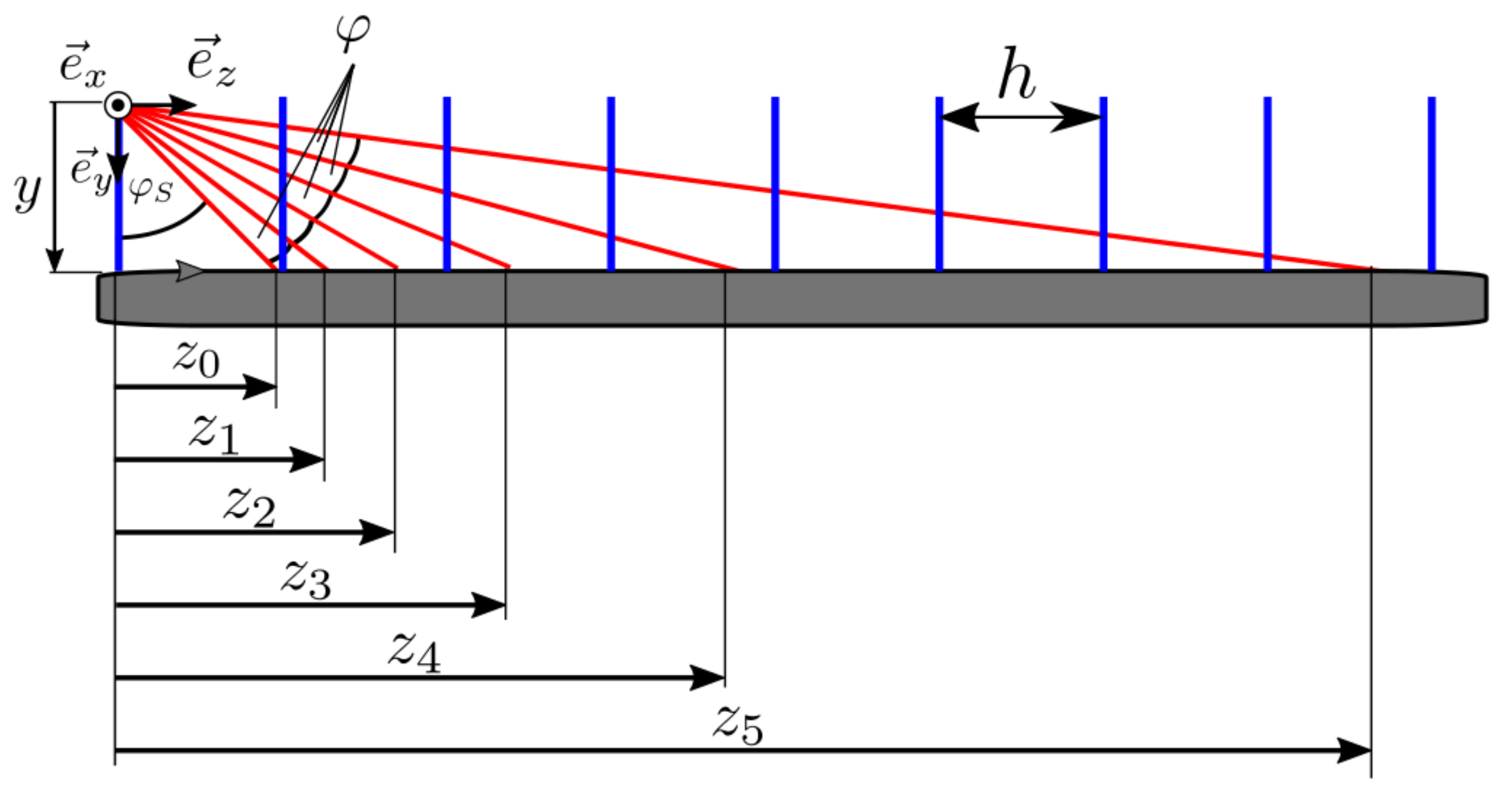}
  \caption{
  The feature points on the ground plane are an important feature especially in highway scenarios.
  Since the depth of the i-th ray is determined by $z_i=y\cdot \tan{(\phi_s+i\cdot\phi)}$, the segmentation with a depth histogram is not applicable.
  Therefore, we detect feature points on the ground plane and fit a local plane without applying the segmentation.}
\label{fig:depth_gp}
\end{figure}

Segmenting the foreground and fitting a local plane is precise for planes orthogonal to the vertical axis of the LIDAR\@.
In contrast, the depth of points on the ground plane cannot be estimated with that method, since the LIDAR has less resolution in vertical than in horizontal direction, as illustrated in Figure~\ref{fig:depth_gp}.
Since feature points on the ground plane are valuable, especially on rural roads and highways, we use a different approach for enabling their depth estimation.
In a first step the ground plane is extracted from the LIDAR point cloud with a robust fit by RANSAC with refinement~\cite{szeliski2010computer}.
Since the road surface is rarely a flat surface, intersecting the viewing ray with the ground plane is inaccurate.
To accurately estimate the depth of $f$ on the road, we rather segment points corresponding to the ground plane than to the foreground.
We estimate local planes around $f$ by the same steps as before, but with a larger minimum spanning area of $F_{\Delta}$.
Outliers can be extracted effectively by accepting only local planes that lie in vicinity to the ground plane.

\section{Block C:\@ Frame to Frame Odometry}
\label{sec:frame2frame}
To obtain an initialization for the Bundle Adjustment, a frame to frame motion estimate is integrated.
Having an accurate frame to frame motion, computation time can be compensated since the prior is typically a much smaller problem to solve and landmark selection with the reconstructed landmarks is more precise.\\
The starting point for our frame to frame motion estimation is the well known Perspective-n-Point-Problem~\cite{szeliski2010computer}
\begin{align}
  & \underset{x,y,z,\alpha,\beta,\gamma}{argmin} \quad \underset{i}{\sum} \| \varphi_{i, 3d\rightarrow2d} \|_2^2 \\
  & \varphi_{3d\rightarrow2d} = \bar{p}_i-\pi(p_i, P(x,y,z,\alpha,\beta,\gamma)),
\end{align}
with $\bar{p}_i$ the measured feature point in the current frame, $p_i$ the 3d-point corresponding to $\bar{p}_i$, $P(x,y,z,\alpha,\beta,\gamma)$ the transform from the previous to the current frame with 3 translative and 3 rotative degrees of freedom and $\pi(\dots)$ the projection function from the 3d world domain to the 2d image domain.
$p_i$ is obtained from the feature correspondence $\tilde{p}_i$ of $\bar{p}_i$ in the previous frame and its estimated depth.\\
This method is well suited for urban scenarios with rich structure.
However, in scenarios with low structure and large optical flow as on highways, the number of extracted features with valid depth can be too small to get a precise estimate.
In order to stabilize the algorithm, we add the epipolar error $\varphi_{2d\rightarrow2d}$~\cite{hartley2003multiple}.
\begin{equation}
  \varphi_{i, 2d\rightarrow2d} = \bar{p}_i F(\frac{x}{z} , \frac{y}{z} ,\alpha,\beta,\gamma) \tilde{p}_i
\end{equation}
with the fundamental matrix F which can be obtained from the frame to frame motion and the intrinsic calibration of the camera.
These additional constraints result also in higher accuracy in rotation.
Additional loss functions are wrapped around the cost functions to reduce the influence of outliers.
We found that the best choice for the loss function is the Cauchy function $\rho_{s}(x)={a(s)}^2\cdot \log{(1+\frac{x}{{a(s)}^2})}$, with the fix outlier threshold $a(s)$. 
For further details on the fundamental matrix and the influence of the loss function, we refer to Hartley et al.~\cite{hartley2003multiple}.
The resulting optimization problem for the frame to frame motion estimate is therefore
\begin{equation}
  \underset{x,y,z,\alpha,\beta,\gamma}{argmin}\quad \underset{i}{\sum} \rho_{3d\rightarrow2d}(\| \varphi_{i, 3d\rightarrow2d} \|_2^2) + \rho_{2d\rightarrow2d}(\| \varphi_{i, 2d\rightarrow2d} \|_2^2).
\end{equation}

\section{Block D:\@ Backend}
\label{sec:backend}
Data accumulation and temporal inference are important steps to increase accuracy and robustness.
Especially for a methodology with sparse depth measurements as presented here, the joint optimization of data collected through many frames makes it less susceptible to errors.
Therefore, we propose a keyframe based Bundle Adjustment framework that is separate in structure and software from the prior estimation.
In the following we describe the key components of the system:
\begin{itemize}
  \item Keyframe selection strategy
  \item Landmark selection strategy
  \item Cost functions
  \item Robustification measures
\end{itemize}
\subsection{Why is Selection Important?}
In an offline process, Bundle Adjustment can be modeled as one big optimization problem, solving for all landmarks and poses jointly.
Having the best accuracy, this comes at a high complexity, disabling the use for online Visual Odometry.
To reduce the computational cost, Bundle Adjustment is solved in optimization windows, removing previous poses and landmarks from the problem.
Since the error is thus minimized locally, drift is larger for the windowed approach than for full Bundle Adjustment. We are in a dilemma: 
On the one hand, the optimization window should be as long as possible to reduce drift, on the other hand the complexity will increase dramatically for longer windows.
In order to obtain the best possible result for low cost, the density of information must be maximized.
Instead of using all data possible, we want to exclude unnecessary measurements while retaining the set that carries the information needed for an accurate pose estimate, which is described in Section~\ref{sec:backend:keyframe_selection} and Section~\ref{sec:backend:landmark_selection}.


\subsection{Keyframe Selection}
\label{sec:backend:keyframe_selection}
The first important step to reduce complexity is the selection of keyframes.
Frames are consecutive instances in time to which incoming data is associated.
A keyframe is a frame that is selected for the online optimization process.
We categorize frames in the following manner:
\begin{itemize}
  \item Required. Frames that must be used for stable optimization.
  \item Rejected. Frames that port invaluable information and should not be used.
  \item Sparsified. Frames that may be used but can be reduced to save computational cost.
\end{itemize}
Required frames port crucial measurements.
Excluding them results in inaccurate and unstable behavior.
This is the case if the viewing angles of the landmarks are changing strongly, for example in turns of the vehicle in dense environment.
The feature matching becomes more difficult and feature tracks become shorter.
To enable overall consistency, a high density of keyframes in turns is needed.
The detection of turns is done by the orientation difference of the frame to frame motion estimate.
Frame rejection becomes important if the vehicle does not move.
If no depth information can be obtained and the image has no optical flow, the problem has one undefined parameter, the scale.
As a result, noise in the feature point detection can lead to incorrect estimation.
Such situations mostly occur at intersections when the ego vehicle stands still and moving vehicles occlude the static environment.
Therefore, no frames will be marked as keyframes, if the mean optical flow is smaller than a fix threshold.
All remaining frames that are neither rejected nor required are gathered and the method chooses frames in time intervals of $0.3 \text{s}$.
As a result the set of keyframes added to the optimization problem is small and comprises a large amount of information to assure the stability and accuracy of the Bundle Adjustment.\\
The last step in the keyframe selection is the choice of the length of the optimization window.
We evaluate the connectivity of the keyframes inside the optimization window by counting landmarks that connect the current keyframe with the newest keyframe.
Non connected keyframes could not contribute to the solution, hence we avoid the waste of computational effort --- if this measure is lower than a threshold, we set the current keyframe as the last keyframe in the optimization window.
To avoid short optimization windows in environments where tracks become short, the size of the optimization window is bounded on both sides.


\subsection{Landmark Selection}
\label{sec:backend:landmark_selection}
\begin{figure}[h]
  \centering
  \def\svgwidth{0.7\columnwidth}
  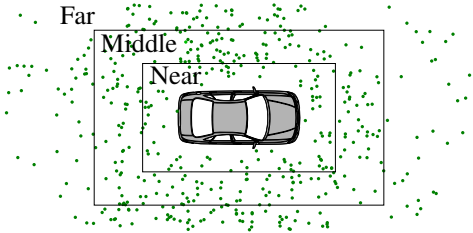%
  \caption{Categorization of landmarks in bins near, middle and far, used for sparsification and selection of landmarks.}
\label{fig:landmark_categories}
\end{figure}

Landmark selection is one of the most discussed topics in the field of Visual Odometry since it is one of the key elements of every visual motion estimation algorithm.
The optimal set of landmarks should satisfy the following conditions:
\begin{itemize}
  \item Well observable, so that the landmarks can be reconstructed precisely.
  \item Small, in order to reduce complexity.
  \item Free of outliers.
  \item Evenly distributed, in 3d space and image space.
\end{itemize}
In this work we aim for late landmark selection.
We match as many feature points as possible with low constraints and filter them afterwards in building block D (Figure~\ref{fig:organigram_global}), when more information can be used for the selection.
By applying the motion estimate, the matches are triangulated, resulting in a sparse point cloud from which landmarks for the Bundle Adjustment are selected.
In a first step, all triangulated feature points are tested for cheirality.
If the match is imprecise, the triangulated point lies behind the image plane and is rejected.\\
The main selection step splits all landmarks in three bins, near, middle and far, each of which has particular importance to the Bundle Adjustment problem:
\begin{itemize}
  \item Near points are important for the translation estimation, but are usually difficult to measure since their optical flow is large.
  \item Middle points are important for both rotation and translation estimation. We use them to recover from local minima.
  \item Far points are important for the rotation estimation and are easier to track, resulting in many measurement for each landmark.
\end{itemize}
This categorization of triangulated points is done in metric space, as illustrated in Figure~\ref{fig:landmark_categories}.


In order to assure evenly distributed landmarks, we apply a voxel filter with median filtering to the 3d landmarks.
In that way, local landmark accumulations such as in trees or bushes are sparsified and do not dominate the Bundle Adjustment.
For each bin a fix number of landmarks is selected for the Bundle Adjustment.
We apply different selection strategies for each bin corresponding to the feature proficiency.
In the near bin, the landmarks belonging to the feature correspondences with the largest optical flow are selected to assure maximum stability in depth for the reconstructed landmark.
If the Bundle Adjustment is stuck in a local minimum, the landmarks and poses converge to a configuration that is locally optimal, but does not correspond to the correct trajectory.
Using before unseen landmarks, instead of optimized landmarks, helps to recover. 
Therefore, we randomly select the landmarks in the middle bin.
Landmarks in the far bin cannot contribute to the translation estimation, whereas they are crucial for rotation estimation.
In order to assure long term stable landmarks and reduce the number of parameters, we choose landmarks with maximum feature point track length for this bin.\\
In a last step semantic information is used to determine the weight of landmarks.
It seems natural that the influence of landmarks on vegetation for Visual Odometry is twofold.
On the one hand, trees have a rich structure, resulting in feature points that are good to track.
On the other hand, vegetation can move, violating the assumption of a static environment.
To examine the influence of landmarks on vegetation, we apply different weight to them than to landmarks on infrastructure in the estimation process.

\subsection{Landmark Depth Insertion}
\label{sec:backend:landmark_depth_insertion}
Without including the depth estimate of the landmarks, the Bundle Adjustment would not be stable, since the scale could not be optimized --- the depth of the landmarks could be altered along with the length of the translation of the vehicle without changing cost in the optimization problem.
To apply scale, the estimated landmark depth must be added to the optimization problem by an additional cost functor $\xi_{i,j}$, which punishes deviation of the landmark depth from the measured depth as formulated in Equation~\ref{equ:depth}.
\begin{equation}
  \xi_{i,j}(l_i, P_j) =
\begin{cases}
  0, \text{\qquad if }l_i \text{ has no depth estimate}\\
  \widehat{d}_{i,j}-\begin{bmatrix} 0 & 0 & 1 \end{bmatrix} \tau(l_i, P_j), \text{\quad else,}  
\end{cases}
\label{equ:depth}
\end{equation}
  
where $l_i$ is the landmark, $\tau$ its mapping from world coordinates to camera coordinates and $\widehat{d}$ is the depth estimate obtained by the method described in Section~\ref{sec:depth_estimation}.
The indices $i,j$ comprise only landmark-pose combinations for which a valid depth estimate could be extracted.
For urban scenarios in which a high number of depth estimates is available, this is a sufficient measure to solve the Bundle Adjustment.
In highway scenarios, however, the Bundle Adjustment has to rely on only a dozen depth estimates that can be error prone.
Hence, additional measures have to be taken.
Here we make use of a popular technique in monocular VSLAM\@.
The oldest motion in the optimization window is in general the most accurate one, since it comprises the maximum information at a given time instance.
We therefore add an additional cost functor $\nu$ that punishes deviations from the length of its translation vector, see Equation~\ref{equ:reg_depth}.
\begin{equation}
  \nu(P_1, P_0)=\widehat{s}(P_1,P_0) - s
\label{equ:reg_depth}
\end{equation}
with $P_0$, $P_1$ the last two poses in the optimization window and $\widehat{s}(P_1,P_0) = \| \text{translation}(P_0^{-1}P_1) \|_2^2$.
$s$ a constant with the value of $\widehat{s}(P_1,P_0)$ before optimization.
In that way changes in scale are regularized and the estimate is smoother and more robust to outliers.

\subsection{Robustification and Problem Formulation}
\label{sec:backend:robustification}
Outliers disable Least-Square methods to converge to the correct minimum (q.v.~\cite{torr2000mlesac},~\cite{torr2004invariant}).
We use semantics and cheirality for a preliminary outlier rejection as mentioned in Section~\ref{sec:backend:landmark_selection}.
Though most of the outliers can be detected by these methods, some still remain such as moving shadows of vehicles, non classified moving objects, etc..
As shown in our previous work~\cite{graeter2017momo}, loss functions improve estimation accuracy and robustness drastically.
Therefore, we employ the Cauchy function as loss functions $\rho_{\phi}(x)$, $\rho_{\xi}(x)$ in order to reduce the influence of large residuals in both, the depth and the reprojection error cost functions.
With $\mathcal{P}$ and $\mathcal{L}$, the sets of keyframes and selected landmarks, the overall optimization problem is hence formulated as
\begin{align}
    & \underset{P_j\in\mathcal{P}, l_i\in\mathcal{L},d_i\in\mathcal{D}}{argmin}\quad w_0\| \nu(P_1, P_0) \|_2^2 \nonumber \\
    & + \underset{i}{\sum}\underset{j}{\sum} w_1\rho_{\phi}( \| \phi_{i,j}(l_i, P_i) \|_2^2 ) + w_2\rho_{\xi}( \| \xi_{i,j}(l_i,P_j) \|_2^2),
\end{align}
with the reprojection error $\phi_{i,j}(l_i,P_j) = \bar{l}_{i,j}-\pi(l_i, P_j)$ and weights $w_0$,$w_1$,$w_2$ that are used to scale the cost functions to the same order of magnitude.
While loss functions reduce the influence of large residuals effectively, the parameters of outlier landmarks remain in the optimization problem and their residuals will be evaluated at each iteration.
Especially for applications with a large number of parameters such as Bundle Adjustment, much computational effort is wasted on outliers.
We propose the use of a trimmed-least-squares-like approach, which removes residuals and parameters after some iterations as shown in Algorithm~\ref{alg:trimmed_least_squares}.

\begin{algorithm}
  \KwData{
    OptimizationProblem p\;
    NumbersSteps ns\;
    RejectionLimit rl\;
  }
  \KwResult{OptimizationProblem\;}
  \ForEach{$s \in ns$}{%
    do $s$ iterations in $p$\;
    remove $rl\%$ highest residuals in depth from $p$\;
    remove $rl\%$ highest residuals in reprojection from $p$\;
    remove all parameters without residuals from $p$\;
  }
  optimize $p$ until convergence or time bound\;
  \caption{Trimmed-least-square-like algorithm for optimization in Bundle Adjustment.
  We use a small set of predefined number of steps to reject residuals in each iteration, improving convergence speed of the optimization problem drastically.
  The final optimization is stopped upon satisfying the convergence criteria or after a fix period of time to assure real time performance.
  }
\label{alg:trimmed_least_squares}
\end{algorithm}

\section{Results and Evaluation}
\label{sec:results_and_evaluation}
\begin{figure}[t]
    \small
    \centering
    \def\svgwidth{\columnwidth}
    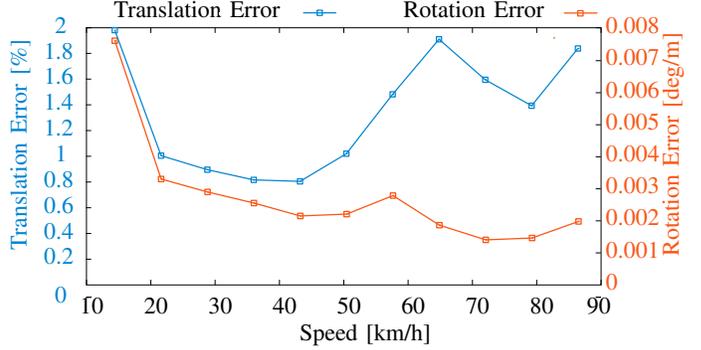%
    \caption{Average errors of the training data set of the KITTI odometry benchmark. 
    Errors are particularly high for high speed and low speed.
    The error at high speed is caused by difficult feature extraction as described in Section~\ref{sec:backend:keyframe_selection}. 
    High error at low speed is caused by conservative tuning of the standstill detection.}
\label{fig:KITTI_results}
\end{figure}
\begin{figure*}[!h]
    \centering
    \small
    \subfloat[Sequence 00]{\def\svgwidth{0.98\columnwidth}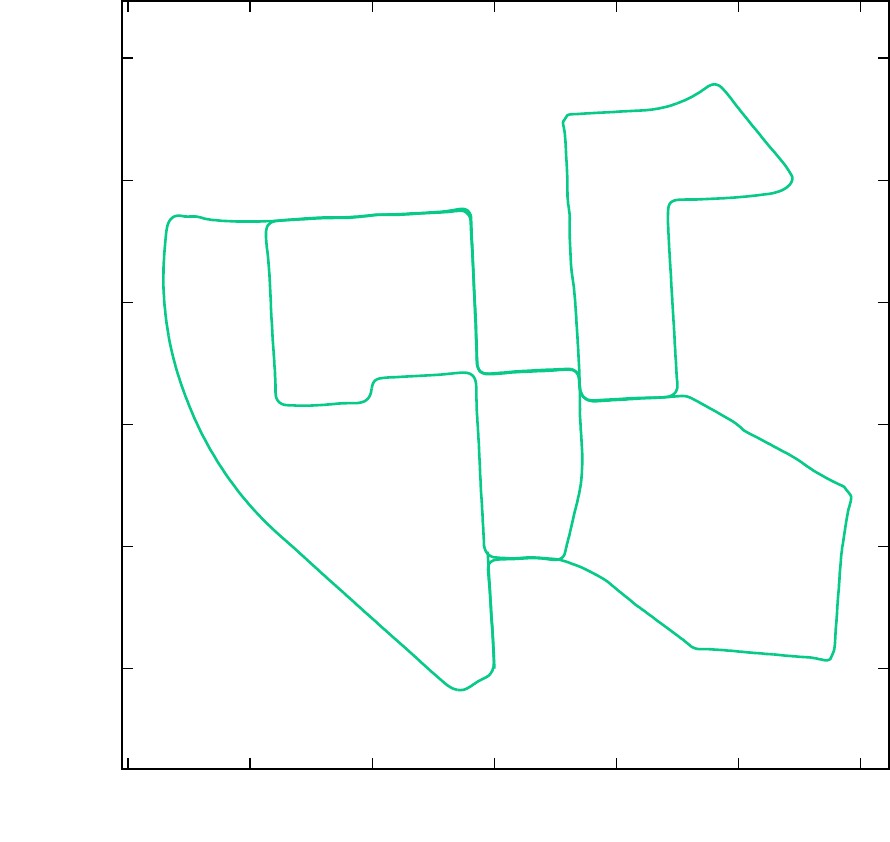%
\label{fig:KITTI_examples:a}}
    \hfill
    \subfloat[Sequence 09]{\def\svgwidth{0.98\columnwidth}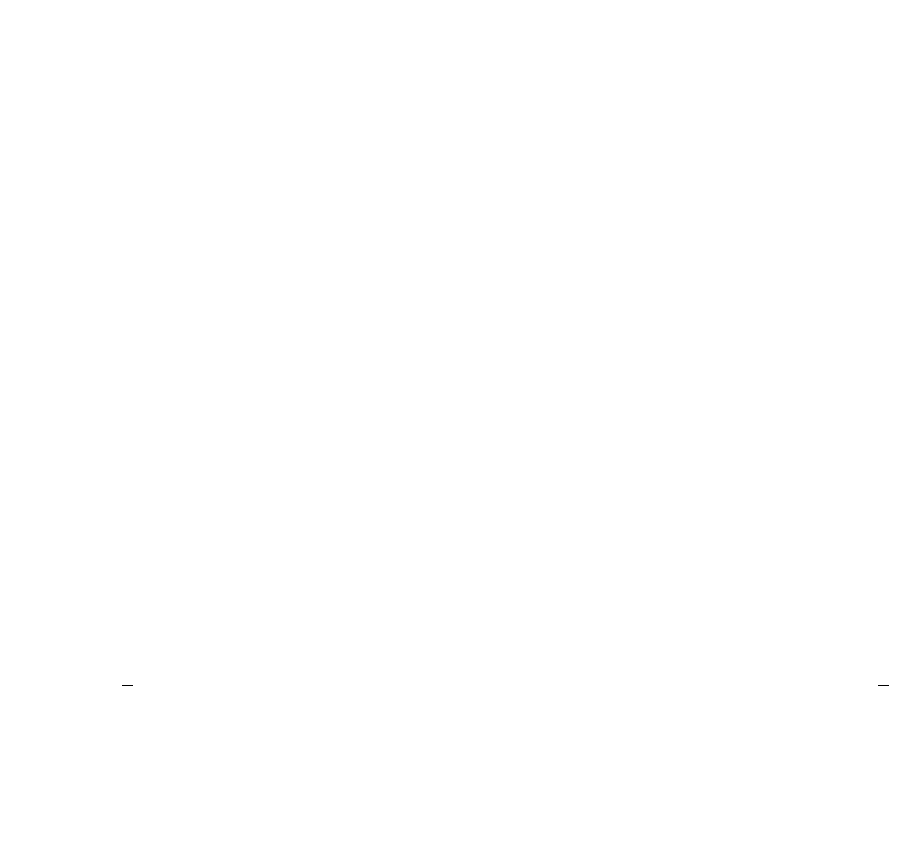%
\label{fig:KITTI_examples:b}}
    \vfill
    \subfloat[Sequence 03]{\def\svgwidth{0.98\columnwidth}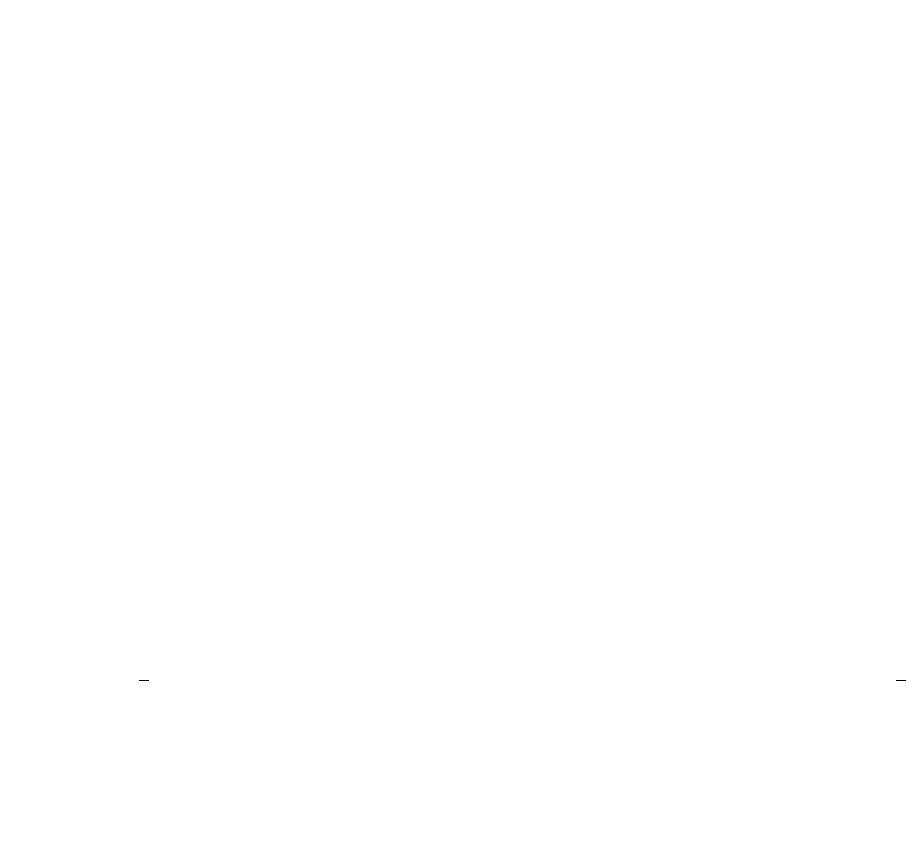%
\label{fig:KITTI_examples:c}}
    \hfill
    \subfloat[Sequence 01]{\def\svgwidth{0.98\columnwidth}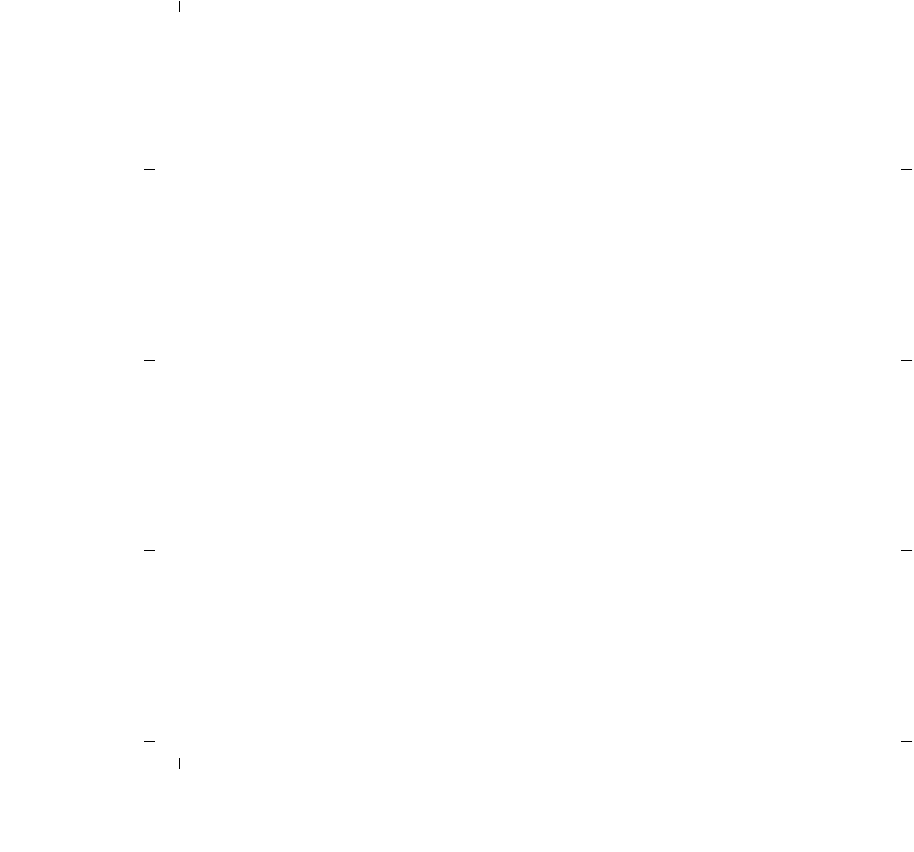%
\label{fig:KITTI_examples:d}}
\caption{Examples of the training set of the KITTI odometry benchmark for the frame to frame motion estimation (Liviodo; blue, light dashed) and the whole pipeline (LIMO\@; black, solid). 
For Visual Odometry on short tracks or with in an environment with a lot of infrastructure (Sequence $00$) Liviodo is suitable.
For longer tracks, the clear benefit of using the backend (Block D) for reducing drift becomes visible. 
In scenes with less infrastructure (Sequence $09$, $03$), the backend additionally compensates imprecise depth estimates caused by violating the local plane assumption.
Using Bundle Adjustment is particularly important on highways such as Sequence $01$, since only few valid depth estimates can be extracted.
}
\label{fig:KITTI_examples}
\end{figure*}
The algorithm has been evaluated on the KITTI dataset which has emerged as the most popular benchmark for Visual Odometry and VSLAM\@.
It includes rural scenery as well as highway sequences and provides gray scale images, color images, LIDAR point clouds and their calibration.
In order to discuss the necessity of the expensive temporal inference block, the frame to frame motion estimation block is evaluated separately from the complete pipeline.


Semantic labels are obtained by a Resnet38 with minor modifications that runs at $100 \text{ms}$ on an Nvidia TitanX Pascal.
It is trained on the Cityscapes dataset~\cite{cordts2016cityscapes}.
The semantic images are binarized and eroded with a kernel size of $21 \text{pixels}$ so that gaps are filled and we avoid taking feature points at object borders.
The label is determined by a majority voting in a $3\times3$ neighborhood of the newest feature point in the track.
Every residual that is added to the problem and is labeled as vegetation is weighted with a fix value.
To evaluate, we run the whole pipeline on the sequences $00$ to $10$ of the KITTI benchmark and compare to the groundtruth using the official KITTI metric~\cite{geiger2013vision}.
We found that a weight of $0.9$ for the vegetation landmarks gives the best results.
However, the optimum vegetation weight varied for different sequences, indicating a more complex relationship between the accuracy of the result and the weight of vegetation landmarks.\\

The pipeline presented here is evaluated on the KITTI dataset~\cite{geiger2013vision}.
Highway scenarios such as sequence $01$, $12$ and $21$ are particularly challenging.
The vehicle drives in open space, so that only the road is in acceptable range for depth estimation.
Since the speed of the vehicle is high, the optical flow on the road is big and motion blur is stronger, making feature tracking difficult.
As a result, only a small amount of valid depth estimates can be extracted.
We resolve this problem by tuning the feature matcher first on the highway scenarios so that the result is stable.
In urban scenarios this, however, causes a large amount of feature matches.
Therefore, the landmark sparsification and selection steps described in Section~\ref{sec:backend:landmark_selection} are of utmost importance.\\
As can be seen in Figure~\ref{fig:KITTI_examples} and Figure~\ref{fig:KITTI_results} the trajectories are precise with very low drift over kilometers of trajectory without using loop closure.
For longer tracks, the benefit of using Bundle Adjustment (Block D) becomes visible --- the drift is reduced drastically.
To demonstrate the applicability of this method for Visual Odometry we evaluate the first pose in the optimization window.
In that way the result is less accurate but available with less latency.
For frames in which no keyframe is chosen, we use the accumulated estimate of the frame to frame motion estimate to extrapolate the pose for evaluation.\\
The results on the evaluation dataset of the frame to frame motion estimation (called Liviodo) and the whole pipeline (called LIMO) are published on the KITTI Visual Odometry benchmark.
Liviodo is placed on rank 30\footnote{\label{ft:date}As of 1st of March 2018.}, with a mean translation error of $1.22\%$ and a rotation error of $0.0042\frac{\deg}{\text{m}}$.
LIMO is placed on rank 13\textsuperscript{\ref{ft:date}}, with a mean translation error of $0.93\%$ and a rotation error of $0.0026\frac{\deg}{\text{m}}$.
Whereas the benefit of using Bundle Adjustment is modest in translation error, the benefit in rotation error of nearly $40\%$ is important.  
This is of particular interest since the rotation error has large influence on the end point error of the trajectory.
LIMO is therefore the second best LIDAR-Camera method published and the best performing method that does not use ICP based LIDAR-SLAM as refinement.
Using a CPU with $3.5 \text{GHz}$, Liviodo runs with $10 \text{Hz}$ on 2 cores and LIMO with $5 \text{Hz}$ on 4 cores.


\section{Conclusions}
\label{sec:conclusion}
Combining highly precise depth measurements from LIDAR and the powerful tracking capabilities of cameras is very promising for Visual Odometry\@.
However, the association of measurements in their specific domains poses an unsolved problem which blocks its application.
In this work, we fill this gap by proposing a methodology for estimating depth from LIDAR for detected features in the camera image by fitting local planes.
Since measurements on the ground plane are particularly important for robust and accurate pose and landmark estimation, they are treated separately.
This is particularly important for highway scenarios.
We embed this methodology in a Bundle Adjustment based Visual Odometry framework by incorporating keyframe selection and landmark selection in order to enable online use.
Our method LIMO is ranked 13th on the competitive KITTI benchmark, outperforming state of the art methods like ORB-SLAM2 and Stereo LSD-SLAM\@.\\
Moreover, we found that giving less weight to vegetation landmarks than to points on infrastructure has a benefit on accuracy.
Therefore, we want to address dynamic landmark weighting in function of semantics and the overall structure of the scene in future research.
We release the code to the community, accessible on Github (https://github.com/johannes-graeter/limo).
\section*{Acknowledgements}
The authors thank the German Research Foundation (DFG) for being funded within the German collaborative research center \textit{SPP 1835 --- Cooperative Interacting Automobiles} (CoInCar).

\IEEEtriggeratref{6}

\bibliographystyle{IEEEtran}
\bibliography{sections/references}

\end{document}